\documentclass{article}

    \usepackage[final]{neurips_2023}

\usepackage[utf8]{inputenc} 
\usepackage[T1]{fontenc}    
\usepackage{hyperref}       
\usepackage{url}            
\usepackage{booktabs}       
\usepackage{amsfonts}       
\usepackage{nicefrac}       
\usepackage{microtype}      
\usepackage{xcolor}         
\usepackage{graphicx}
\usepackage{amsmath}
\usepackage{amssymb}
\usepackage{booktabs}
\usepackage{algorithm}
\usepackage{algpseudocode}
\newcommand{\cmark}{\ding{51}}%
\newcommand{\xmark}{\ding{55}}%

\usepackage[utf8]{inputenc} 
\usepackage[T1]{fontenc}    
\usepackage{hyperref}       
\usepackage{url}            
\usepackage{booktabs}       
\usepackage{amsfonts}       
\usepackage{nicefrac}       
\usepackage{microtype}      
\usepackage{amsmath}
\usepackage{rotating}
\usepackage{amssymb}
\usepackage{pifont}

\title{Turn Down the Noise: Leveraging Diffusion Models for Test-time Adaptation via Pseudo-label Ensembling}

\author{%
  Mrigank Raman \\
  Machine Learning Department\\
  Carnegie Mellon University\\
  \texttt{mrigankr@cmu.edu} \\
  \And
   Rohan Shah \\
  Machine Learning Department\\
  Carnegie Mellon University\\
  \texttt{rohans@cmu.edu} \\
  \And
  Akash Kannan \\
  Machine Learning Department\\
  Carnegie Mellon University\\
  \texttt{akashkan@cmu.edu} \\
  \And
  Pranit Chawla \\
  Machine Learning Department\\
  Carnegie Mellon University\\
  \texttt{pranitc@cmu.edu} \\
}

\begin{document}

\maketitle

\begin{abstract}
  The goal of test-time adaptation is to adapt a source-pretrained model to a continuously changing target domain without relying on any source data. Typically, this is either done by updating the parameters of the model (model adaptation) using inputs from the target domain or by modifying the inputs themselves (input adaptation). However, methods that modify the model suffer from the issue of compounding noisy updates whereas methods that modify the input need to adapt to every new data point from scratch while also struggling with certain domain shifts. We introduce an approach that leverages a pre-trained diffusion model to project the target domain images closer to the source domain and iteratively updates the model via pseudo-label ensembling. Our method combines the advantages of model and input adaptations while mitigating their shortcomings. Our experiments on CIFAR-10C demonstrate the superiority of our approach, outperforming the strongest baseline by an average of $1.7\%$ across 15 diverse corruptions and surpassing the strongest input adaptation baseline by an average of $18\%$. 
\end{abstract}

\section{Introduction}
Supervised deep learning algorithms often assume that the data used for training (source) and testing (target) follow the same ideal pattern, where they are independent and identically distributed (IID). However, in real-world situations, this ideal rarely holds true. Even small differences or changes in data distribution can seriously affect the performance of these models~\citep{QuioneroCandela2009DatasetSI}. This becomes a major challenge when using supervised learning for tasks that involve data with different distributions. To deal with shifts in data distribution, adaptation methods typically use the target data to continuously improve and update their predictions.

Typically, adaptation techniques combat the problem of distribution shift by jointly optimizing on the source and the target distributions during training~\citep{ganin2016domain, hoffman2018cycada, 10.1007/978-3-642-15561-1_16}. These methods exhibit strong performance when the shifts in the target domain are well-understood and predefined. However, they tend to falter when faced with the arrival of an entirely new and previously unseen target domain. This underscores the imperative need for development of test-time adaptation, a technique that adapts the model at inference time without relying on the source data and without interrupting the inference process. Broadly speaking, there are two orthogonal ways of performing test-time adaptation. First, model adaptation~\citep{wang2022continual, wang2020tent, chen2022contrastive, liang2020shot} wherein the model is updated iteratively during inference without using test labels. Second, Input adaptation~\citep{gao2022back} wherein the input is modified to match the source distribution. Model adaptation helps to continually adapt to changing distributions but may suffer from noisy updates depending on the amount of distribution shift. Moreover, input adaptation must adapt to each example from scratch and sometimes struggle with certain distribution shifts. 

Our proposed method, D-TAPE (Diffusion-infused Test-time Adaptation via Pseudo-label Ensembling), seamlessly integrates the strengths of model and input adaptation while minimizing their weaknesses. Utilizing a source domain-trained diffusion model, we align target images more closely with the source domain. However, a direct application of a pre-trained diffusion model risks losing class information. To rectify this, we implement a low-pass filtering technique, akin to \citep{choi2021ilvr}, enabling class-preserving projections. Our model then iteratively updates by ensembling pseudo-labels from both the projected and original test images, offering dynamic adaptability to sudden domain shifts. D-TAPE outperforms existing state-of-the-art approaches like CoTTA ~\citep{wang2022continual}, TENT~\citep{wang2020tent}, and DDA~\citep{gao2022back}. We also illustrate how D-TAPE effectively brings the input data closer to the source domain compared to the corrupted images, as evidenced by a reduction in the $\mathcal{A}$-distance~\citep{a-dist}. We anticipate that this study will serve as a catalyst for the broader adoption and exploration of diffusion models in future test-time adaptation research.

\section{Prior Work}
\label{sec:formatting}

\paragraph{Test Time Adaptation:}
The goal of test-time adaptation is to adapt a source-pretrained model to a continuously changing target domain without relying on any source data. To that end, the landscape of Test Time Adaptation can broadly be segmented into two paradigms namely Model Adaptation and Input Adaptation. In the realm of Model Adaptation, particularly in source-free and test-time settings, various methodologies have emerged. For instance, models like TENT~\citep{wang2020tent}, SHOT~\citep{liang2020shot}, and MEMO~\citep{zhang2021memo} employ unsupervised loss minimization techniques specifically tailored to target data distributions. Alternatively, some strategies resort to the generation of pseudo labels via a range of innovative techniques such as a conjugate pseudo-label function~\citep{goyal2022test}, weight-averaged teacher models~\citep{wang2022continual}, and nearest neighbor soft-voting mechanisms~\citep{chen2022contrastive}. These pseudo-labels serve as scaffolding for self-training procedures, effectively amplifying the model's generalization capabilities. In contrast, Input Adaptation strategies predominantly leave the model untouched and only update the input examples from the target domain. Among these, Diffusion Driven Adaptation~\citep{gao2022back} stands out for its application of a diffusion model trained on source data, thereby enabling more effective test-time adaptation.

\paragraph{Diffusion Modelling:}
Diffusion modeling serves as a pivotal approach in generative techniques, relying on a two-step process that initially introduces noise into data, followed by a denoising phase. Seminal works such as DDPM~\citep{ho2020denoising} and DDIM~\citep{song2020denoising} have been instrumental in establishing the framework of this field. An advanced extension is Guided Diffusion~\citep{dhariwal2021diffusion}, which leverages class labels to enhance the quality of generated images. Since we do not have access to test labels, we employ DDPM models that have been specifically trained on the CIFAR10 dataset for projecting target images closer to the source domain.

\section{Methodology}

\paragraph{Diffusion with Interpolated Latent Refinement: } Initially, a diffusion model is pre-trained to generate images from the source distribution. Then, at test-time, we run the forward process on the domain-shifted input image by iteratively adding Gaussian noise $N$ times, resulting in a sequence $x_0,\ldots,x_N$. We subsequently apply a modified reverse process to project the image closer to the source domain, using the diffusion model trained to maximize the likelihood of the source data. Initially, we sample $\hat{x}^g_{N-1} \sim p_\theta (x^g_{N-1}|x^g_N=x_N)$ from the reverse denoising process of the diffusion model. Since the diffusion model is not explicitly conditioned to preserve class information, we apply the low-pass filter $h()$ from ILVR \citep{choi2021ilvr} to the sampled $\hat{x}^g_{N-1}$ to constrain the low-frequency features of $\hat{x}^g_{N-1}$ to be equal to that of $x_{N-1}$ in the same manner as \cite{gao2022back}. This is a sequence of downsampling and upsampling operations that helps preserve the high-level image structure and hence the class information. However, downsampling results in blurring for low-resolution images. To mitigate this effect, we interpolate the filtered image $h(\hat{x}^g_{N-1})$ with the original sample $\hat{x}^g_{N-1}$ to generate $x^g_{N-1}$. This is fed back into the diffusion model and the process is repeated to generate $x^g_{N-2}$ and so on. We provide the exact algorithm below.

\paragraph{Model Adaptation via Pseudo-label ensembling:}Under the self-training framework, the model weights are adapted using some form of the model predictions themselves. Given a pre-trained model, $f_\theta()$ we follow a strategy similar to that used in CoTTA \citep{wang2022continual} to adapt the model weights. We store two copies of the original model - a student $f_{\theta_s}()$ and a teacher $f_{\theta_t}()$. The teacher generates pseudo-labels that provide supervision to update the student model. The weights of the teacher themselves are adapted by exponential moving average using the student weights. We incorporate the diffusion model output $x^g_0$ in both student predictions $\hat{y}_S = g(x_0, x^g_0, \theta_S)$ and pseudo-labels $\hat{y}_T = g(x_0, x^g_0, \theta_T)$  where the function $g$ is a combination of a conditional ensembling scheme and logit averaging. We provide ablations for these choices in Table~\ref{tab:ablation}. Mathematically,

\[
g(x_0, x^g_0, \theta) = 
\begin{cases} 
f_{\theta}(x_0) & \text{if } \max_{c}f_{\theta_T}(x_0) > \max_{c}f_{\theta_T}(x_0^g)\\ \\
\frac{f_{\theta}(x_0) + f_{\theta_T}(x_0^g)}{2} & \text{otherwise}
\end{cases}
\]

We additionally generate student predictions on augmented versions of the inputs $x'_0$ and $x'^g_0$, and denote $\hat{y}_S' = g(x'_0, x'^{g}_0, \theta_S)$. The loss used to update the student is $\mathcal{L} = -\sum\limits_c 0.5 {y_{T}^c} (\log \hat{y}_{S}^c + \log\hat{y}_{S}^{'c})$.






\begin{figure}
    \centering
    \includegraphics[width=0.7\linewidth]{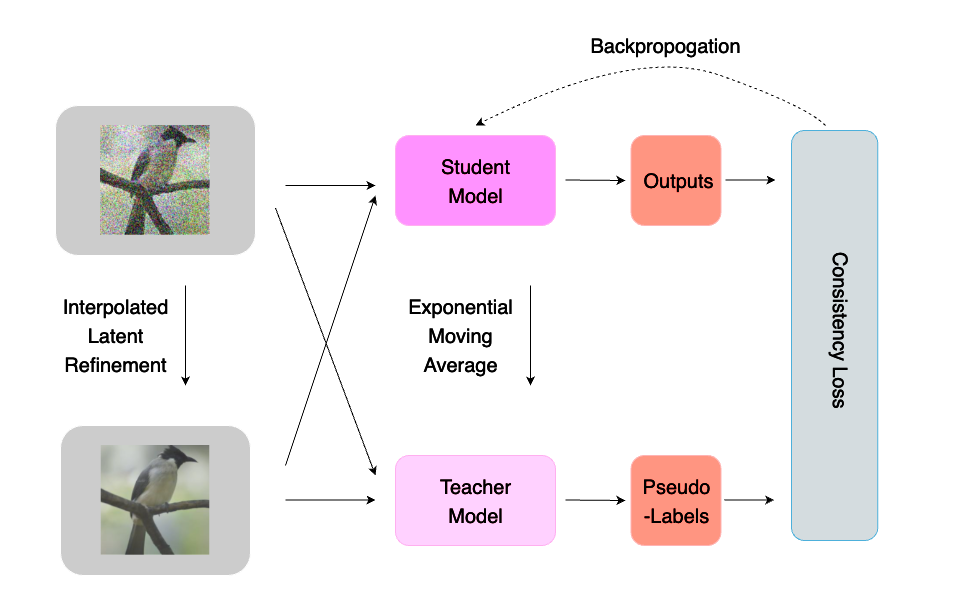}
    \caption{Illustration of our complete adaptation mechanism. Given an image from the target distribution, we first apply our interpolated latent refinement procedure to project the image closer to the source distribution. Then we use a student-teacher framework and update the models by using an ensemble of pseudo-labels generated by the teacher using both the original and projected test image.}
    \label{fig:model_diag}
\end{figure}\begin{algorithm}
\label{main_algo}
\caption{Generation using Diffusion}
\begin{algorithmic}[1] 
\State \textbf{Input:} Reference image $x_0$
\State \textbf{Output:} Generated image $x^{g}_{0}$
\State $N$: diffusion range, $\phi_D(\cdot)$: low-pass filter of scale $D$
\State Sample $x_N \sim q(x_N|x_0)$ \Comment{perturb input}
\State $x^{g}_{N} \leftarrow x_N$
\For{$t \leftarrow N$ \textbf{to} $1$}
    \State $\hat{x}^{g}_{t-1} \sim p_{\theta}(x^{g}_{t-1}|x^{g}_{t})$ \Comment{unconditional proposal}
    \State $\hat{x}^{g}_{0} \leftarrow \frac{1}{\bar{\alpha}_t} x^{g}_{t} - \frac{1}{\bar{\alpha}_t - 1}\epsilon_{\theta}(x^{g}_{t},t)$
    \State $x^{g}_{t-1} \leftarrow \hat{x}^{g}_{t-1} - w \nabla_{xt} \lVert \phi_D(x_0) - \phi_D(\hat{x}^{g}_{0}) \rVert^2$
    \State  $x^{g}_{t-1} \leftarrow \alpha x^{g}_{t-1} + (1-\alpha) \hat{x}^g_{t-1}$ \color{black}
\EndFor
\State \Return $x_{0}^g$
\end{algorithmic}
\end{algorithm}

\section{Experimental Setup and Results}
\paragraph{Datasets and Evaluation Methodology:}
We train our models on the CIFAR10 \citep{krizhevsky2009learning} train set and evaluated on CIFAR10-C \citep{hendrycks2019robustness}. CIFAR10-C comprises of 15 different corruptions at 5 different severity levels.
We evaluate the continuous adaptability of the models using the approach used in \citep{wang2022continual, gan2022decorate} which can briefly described as follows: 
We arrange the corruption types in the same order as demonstrated in Table~\ref{tab:sudden_cifar} and perform inference on the model using batches of test images with a severity level of 5 for each corruption type. This process results in inference on 15 times the size of the clean test data. We exclusively use images with a severity level of 5 to introduce sudden distribution shifts in the batches. We report error rates for each corruption type and calculate the mean error rate across all 15 corruptions. We also provide details and results on a different setting namely gradual TTA in \ref{tab:cifar_grad} 


\paragraph{Baselines:}
We compare D-TAPE to the following test-time adaptation methods. \\ 
\textbf{TENT}~\citep{wang2020tent}: Minimizes model prediction entropy on test data using the model's own predictions as pseudo labels, updating only normalization parameters.\\
\textbf{CoTTA}~\citep{wang2022continual}: Utilizes a weight-averaged teacher model for pseudo-label generation and updates a student network. It also incorporates augmentation-averaged pseudo labels for low-confidence test inputs and restores a fraction of model weights to their original values.\\
 \textbf{AdaContrast}~\citep{chen2022contrastive}: Maintains a memory queue of target features and predictions with a momentum model. Generates pseudo-labels for using a combination of cross-entropy loss and contrastive loss to update the model.\\
 \textbf{DDA}~ \citep{gao2022back}: Employs a diffusion process to transform test images back to the source domain, using the output as predictions. No parameter updates occur during testing, and a low-frequency filter retains class information during diffusion. Following the approach of \cite{wang2022continual}, we employ a pre-trained WideResNet-28-10~\citep{zagoruyko2016wide} from RobustBench~\citep{croce2020robustbench} as the backbone for all baselines on CIFAR10-C.

\label{sec:baselines}
 

\paragraph{Results Discussion:}
Our method outperforms the strongest baseline which is CoTTA on $14$ out of the $15$ corruptions while beating it by an average of $1.7\%$ across all $15$ corruptions ~(Table \ref{tab:sudden_cifar}). Additionally, our method also beats DDA which only modifies the input by an average of $18\%$ across all corruptions. This underscores the effectiveness of our method.  
\begin{table*}[ht]
\centering
\scalebox{0.78}{
\setlength\tabcolsep{2pt}
\begin{tabular}{@{}c|cccccccccccccccc@{}}
\toprule
Method      & \begin{turn}{60}gauss \end{turn}& \begin{turn}{60}shot \end{turn} & \begin{turn}{60}impulse \end{turn}& \begin{turn} {60} defocus \end{turn} & \begin{turn} {60} glass \end{turn} & \begin{turn} {60} motion \end{turn} & \begin{turn} {60} zoom \end{turn} & \begin{turn} {60} snow \end{turn}  & \begin{turn} {60} frost \end{turn} & \begin{turn} {60}fog \end{turn}   & \begin{turn} {60}brightness \end{turn}& \begin{turn} {60} contrast \end{turn}& \begin{turn} {60} elastic \end{turn} & \begin{turn} {60} pixelate \end{turn} & \begin{turn} {60} jpeg \end{turn} & \begin{turn} {60} mean \end{turn} \\ \midrule
Source      & 72.33 & 65.71 & 72.92   & 46.95   & 54.33 & 34.76  & 42.01 & 25.09 & 41.30 & 26.01 & 9.31       & 46.71    & 26.59   & 58.45    & 30.30 & 43.52 \\
DDA         & 33.09 & 30.38 & 38.18   & 49.51   & 33.61 & 37.94  & 41.28 & 19.70 & 24.18 & 39.10 & 11.22      & 61.24    & 25.38   & 30.95    & 19.47 & 33.02 \\
TENT        & 24.79 & 20.48 & 28.54   & 14.54   & 32.27 & 16.05  & 13.88 & 20.21 & 20.45 & 17.77 & 11.17      & 15.20    & 24.06   & 19.96    & 25.38 & 20.32 \\
AdaContrast & 29.19 & 22.50 & 30.09   & 13.85   & 32.82 & 14.00  & 12.10 & 16.47 & 14.73 & 14.31 & 8.04       & \textbf{9.83}     & 22.11   & 17.76    & 19.95 & 18.52 \\
COTTA       & 24.30 & 21.43 & 25.91   & 11.72   & 27.90 & 12.33  & 10.66 & 14.84 & 13.82 & 12.36 & \textbf{7.54}       & 10.76    & 18.23   & 13.61    & 17.68 & 16.21 \\
D-TAPE~(Ours)        & \textbf{19.29} & \textbf{17.19} & \textbf{22.58}   & \textbf{11.54}   & \textbf{22.73} & \textbf{11.94}  & \textbf{10.05} & \textbf{14.13} & \textbf{12.98} & \textbf{12.15} & 8.17       & 10.58    & \textbf{16.87}   & \textbf{12.70}    & \textbf{14.54} & \textbf{14.50} \\ 
\bottomrule 
\end{tabular}
}
\vspace{0.3pt}
\caption{Classification error rate (\%) for the standard CIFAR10-to-CIFAR10-C time adaptation task in the sudden setting. Evaluated on WideResNet-28 with the severity level 5.}
\label{tab:sudden_cifar}
\end{table*}

\paragraph{Gradual TTA: }Unlike the sudden setting, here we make use of all the severity levels. Like the sudden setting,  we generate a permutation of all the different types of corruptions and for each type we sample batches with severity in the following order $1\rightarrow2\rightarrow3\rightarrow4\rightarrow5\rightarrow4\rightarrow3\rightarrow2\rightarrow1$ on which we do inference, update weights and then go to the next type of corruption in the same ordering of severity. Note that we do not reset the weights to the pre-trained values anywhere in between. For each type of corruption, the average error on its images is the indicator of performance. D-TAPE outperforms the strongest baseline by $\sim 1\%$ on average across $15$ corruptions and improves massively over the best input adaptation technique which is DDA~(Table~\ref{tab:cifar_grad}).

 \begin{table*}[ht]
\centering
\begin{tabular}{@{}cc@{}}
\toprule
Model       & Gradual CIFAR-10C\\ \midrule
Source      & 43.52                                      \\
DDA      & 33.02                                  \\
TENT      & 24.56                                \\
AdaContrast & 15.78                                     \\
CoTTA   & 13.99                       \\
D-TAPE   & \textbf{13.09}                       \\ \bottomrule
\end{tabular}
\caption{Mean results over different types of corruption over the gradual CIFAR10-C at level 5 corruption, we report error percentage for each method. }
\label{tab:cifar_grad}
\end{table*}

\begin{figure}[ht]
    \centering
    \includegraphics[width=0.7\linewidth]{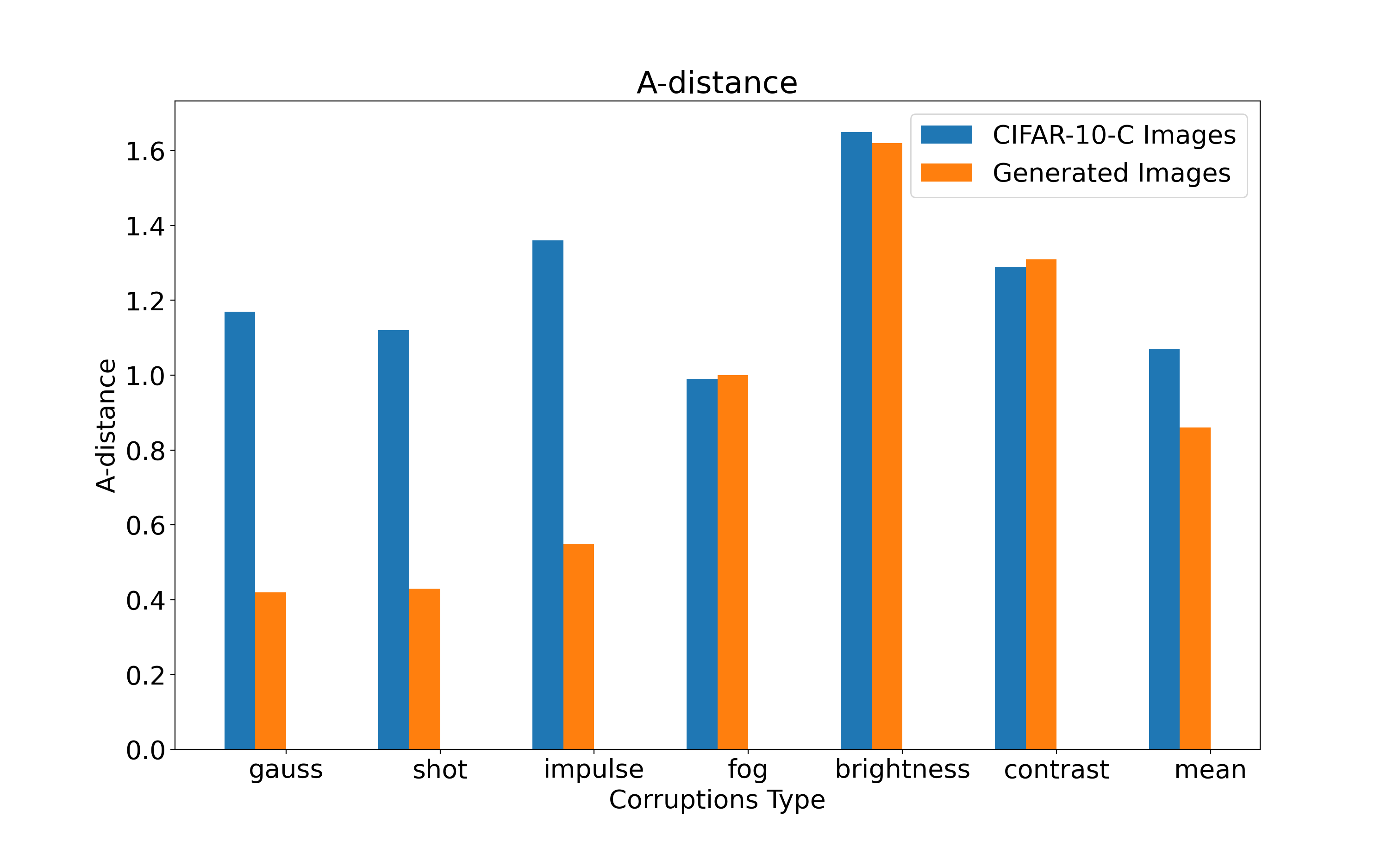}
    \caption{$\mathcal{A}$-distance between the CIFAR-10C dataset and the original CIFAR-10 dataset, and $\mathcal{A}$-distance between generated images and original CIFAR-10 dataset. Mean is over 15 noises}
    \label{fig:adis}
\end{figure}

\paragraph{Ablations}
We perform ablations to demonstrate the importance of using conditional ensembling and logit averaging in the pseudo-label ensembling phase of D-TAPE and report the results in Table~\ref{tab:ablation}. We notice that without conditional ensembling, our performance on fog and contrast drastically decreases. This is expected since diffusion fails to project the images closer to the source for these corruptions as seen in Figure~\ref{fig:adis}. We also observe that without logit averaging our performance drops slightly in the case of noises as well where we expect the best performance. Logit averaging helps us to defend against the rare cases where diffusion fails to retain the same object in the image.

\begin{table*}[ht]
\centering
\begin{tabular}{@{}ll|llllll@{}}
\toprule
Conditional Ensembling & Logit Averaging & gauss & shot  & impulse & fog   & contrast & mean  \\ \midrule
\xmark                & \xmark           & 26.19 & 24.9  & 33.76   & 33.69 & 41.19    & 23.68 \\
\xmark                      & \cmark              & 19.70  & 18.10  & 23.79   & 16.79 & 17.37    & 16.02 \\
        \cmark         & \xmark           & 20.21 & 18.57 & 24.26   & 14.73 & 14.14    & 16.03 \\
\cmark                    & \cmark           & \textbf{19.29} & \textbf{17.19} & \textbf{22.58}   & \textbf{12.15} & \textbf{10.58} &   \textbf{14.50} \\ \bottomrule
\end{tabular}
\caption{Ablation on two components of our model, conditional ensembling and logit averaging. Mean is over 15 noises}
\label{tab:ablation}
\end{table*}

\paragraph{Does Diffusion help? }To answer the question of whether using diffusion helps or not, we use $\mathcal{A}$-distance~\citep{a-dist} to quantify shifts in distribution. We follow \citet{a-dist} and train a linear classifier to separate the two domains for computing the $\mathcal{A}$-distance. We observe that, on average, the generated images closely match the clean image distribution, with particularly close alignment in the case of gaussian, shot, and impulse noises. However, for fog and contrast, the $\mathcal{A}$-distance slightly increases for generated images.(Figure~\ref{fig:adis}). We believe this happens because our interpolated latent refinement scheme retains low frequency noises. We report all the $\mathcal{A}$-distance values in Table~\ref{app:tab:A-distance} which indicate that diffusion is beneficial for almost all corruptions.

\bibliography{references}

\begin{thebibliography}{21}
\providecommand{\natexlab}[1]{#1}
\providecommand{\url}[1]{\texttt{#1}}
\expandafter\ifx\csname urlstyle\endcsname\relax
  \providecommand{\doi}[1]{doi: #1}\else
  \providecommand{\doi}{doi: \begingroup \urlstyle{rm}\Url}\fi

\bibitem[Quionero-Candela et~al.(2009)Quionero-Candela, Sugiyama, Schwaighofer, and Lawrence]{QuioneroCandela2009DatasetSI}
Joaquin Quionero-Candela, Masashi Sugiyama, Anton Schwaighofer, and Neil~D. Lawrence.
\newblock Dataset shift in machine learning.
\newblock 2009.
\newblock URL \url{https://api.semanticscholar.org/CorpusID:61294087}.

\bibitem[Ganin et~al.(2016)Ganin, Ustinova, Ajakan, Germain, Larochelle, Laviolette, Marchand, and Lempitsky]{ganin2016domain}
Yaroslav Ganin, Evgeniya Ustinova, Hana Ajakan, Pascal Germain, Hugo Larochelle, Fran{\c{c}}ois Laviolette, Mario Marchand, and Victor Lempitsky.
\newblock Domain-adversarial training of neural networks.
\newblock \emph{The journal of machine learning research}, 17\penalty0 (1):\penalty0 2096--2030, 2016.

\bibitem[Hoffman et~al.(2018)Hoffman, Tzeng, Park, Zhu, Isola, Saenko, Efros, and Darrell]{hoffman2018cycada}
Judy Hoffman, Eric Tzeng, Taesung Park, Jun-Yan Zhu, Phillip Isola, Kate Saenko, Alexei Efros, and Trevor Darrell.
\newblock Cycada: Cycle-consistent adversarial domain adaptation.
\newblock In \emph{International conference on machine learning}, pages 1989--1998. Pmlr, 2018.

\bibitem[Saenko et~al.(2010)Saenko, Kulis, Fritz, and Darrell]{10.1007/978-3-642-15561-1_16}
Kate Saenko, Brian Kulis, Mario Fritz, and Trevor Darrell.
\newblock Adapting visual category models to new domains.
\newblock In Kostas Daniilidis, Petros Maragos, and Nikos Paragios, editors, \emph{Computer Vision -- ECCV 2010}, pages 213--226, Berlin, Heidelberg, 2010. Springer Berlin Heidelberg.
\newblock ISBN 978-3-642-15561-1.

\bibitem[Wang et~al.(2022)Wang, Fink, Van~Gool, and Dai]{wang2022continual}
Qin Wang, Olga Fink, Luc Van~Gool, and Dengxin Dai.
\newblock Continual test-time domain adaptation.
\newblock In \emph{Proceedings of the IEEE/CVF Conference on Computer Vision and Pattern Recognition}, pages 7201--7211, 2022.

\bibitem[Wang et~al.(2020)Wang, Shelhamer, Liu, Olshausen, and Darrell]{wang2020tent}
Dequan Wang, Evan Shelhamer, Shaoteng Liu, Bruno Olshausen, and Trevor Darrell.
\newblock Tent: Fully test-time adaptation by entropy minimization, 2020.
\newblock URL \url{https://arxiv.org/abs/2006.10726}.

\bibitem[Chen et~al.(2022)Chen, Wang, Darrell, and Ebrahimi]{chen2022contrastive}
Dian Chen, Dequan Wang, Trevor Darrell, and Sayna Ebrahimi.
\newblock Contrastive test-time adaptation.
\newblock In \emph{Proceedings of the IEEE/CVF Conference on Computer Vision and Pattern Recognition}, pages 295--305, 2022.

\bibitem[Liang et~al.(2020)Liang, Hu, and Feng]{liang2020shot}
Jian Liang, Dapeng Hu, and Jiashi Feng.
\newblock Do we really need to access the source data? source hypothesis transfer for unsupervised domain adaptation, 2020.
\newblock URL \url{https://arxiv.org/abs/2002.08546}.

\bibitem[Gao et~al.(2022)Gao, Zhang, Liu, Darrell, Shelhamer, and Wang]{gao2022back}
Jin Gao, Jialing Zhang, Xihui Liu, Trevor Darrell, Evan Shelhamer, and Dequan Wang.
\newblock Back to the source: Diffusion-driven test-time adaptation.
\newblock \emph{arXiv preprint arXiv:2207.03442}, 2022.

\bibitem[Choi et~al.(2021)Choi, Kim, Jeong, Gwon, and Yoon]{choi2021ilvr}
Jooyoung Choi, Sungwon Kim, Yonghyun Jeong, Youngjune Gwon, and Sungroh Yoon.
\newblock Ilvr: Conditioning method for denoising diffusion probabilistic models, 2021.

\bibitem[Ben-David et~al.(2006)Ben-David, Blitzer, Crammer, and Pereira]{a-dist}
Shai Ben-David, John Blitzer, Koby Crammer, and Fernando Pereira.
\newblock Analysis of representations for domain adaptation.
\newblock In B.~Sch\"{o}lkopf, J.~Platt, and T.~Hoffman, editors, \emph{Advances in Neural Information Processing Systems}, volume~19. MIT Press, 2006.
\newblock URL \url{https://proceedings.neurips.cc/paper_files/paper/2006/file/b1b0432ceafb0ce714426e9114852ac7-Paper.pdf}.

\bibitem[Zhang et~al.(2021)Zhang, Levine, and Finn]{zhang2021memo}
Marvin Zhang, Sergey Levine, and Chelsea Finn.
\newblock Memo: Test time robustness via adaptation and augmentation, 2021.
\newblock URL \url{https://arxiv.org/abs/2110.09506}.

\bibitem[Goyal et~al.(2022)Goyal, Sun, Raghunathan, and Kolter]{goyal2022test}
Sachin Goyal, Mingjie Sun, Aditi Raghunathan, and J~Zico Kolter.
\newblock Test time adaptation via conjugate pseudo-labels.
\newblock In Alice~H. Oh, Alekh Agarwal, Danielle Belgrave, and Kyunghyun Cho, editors, \emph{Advances in Neural Information Processing Systems}, 2022.
\newblock URL \url{https://openreview.net/forum?id=2yvUYc-YNUH}.

\bibitem[Ho et~al.(2020)Ho, Jain, and Abbeel]{ho2020denoising}
Jonathan Ho, Ajay Jain, and Pieter Abbeel.
\newblock Denoising diffusion probabilistic models.
\newblock \emph{Advances in Neural Information Processing Systems}, 33:\penalty0 6840--6851, 2020.

\bibitem[Song et~al.(2020)Song, Meng, and Ermon]{song2020denoising}
Jiaming Song, Chenlin Meng, and Stefano Ermon.
\newblock Denoising diffusion implicit models.
\newblock \emph{arXiv preprint arXiv:2010.02502}, 2020.

\bibitem[Dhariwal and Nichol(2021)]{dhariwal2021diffusion}
Prafulla Dhariwal and Alexander Nichol.
\newblock Diffusion models beat gans on image synthesis.
\newblock \emph{Advances in Neural Information Processing Systems}, 34:\penalty0 8780--8794, 2021.

\bibitem[Krizhevsky et~al.(2009)Krizhevsky, Hinton, et~al.]{krizhevsky2009learning}
Alex Krizhevsky, Geoffrey Hinton, et~al.
\newblock Learning multiple layers of features from tiny images.
\newblock 2009.

\bibitem[Hendrycks and Dietterich(2019)]{hendrycks2019robustness}
Dan Hendrycks and Thomas Dietterich.
\newblock Benchmarking neural network robustness to common corruptions and perturbations.
\newblock \emph{Proceedings of the International Conference on Learning Representations}, 2019.

\bibitem[Gan et~al.(2022)Gan, Ma, Lou, Bai, Zhang, Shi, and Luo]{gan2022decorate}
Yulu Gan, Xianzheng Ma, Yihang Lou, Yan Bai, Renrui Zhang, Nian Shi, and Lin Luo.
\newblock Decorate the newcomers: Visual domain prompt for continual test time adaptation.
\newblock \emph{arXiv preprint arXiv:2212.04145}, 2022.

\bibitem[Zagoruyko and Komodakis(2016)]{zagoruyko2016wide}
Sergey Zagoruyko and Nikos Komodakis.
\newblock Wide residual networks.
\newblock \emph{arXiv preprint arXiv:1605.07146}, 2016.

\bibitem[Croce et~al.(2020)Croce, Andriushchenko, Sehwag, Debenedetti, Flammarion, Chiang, Mittal, and Hein]{croce2020robustbench}
Francesco Croce, Maksym Andriushchenko, Vikash Sehwag, Edoardo Debenedetti, Nicolas Flammarion, Mung Chiang, Prateek Mittal, and Matthias Hein.
\newblock Robustbench: a standardized adversarial robustness benchmark.
\newblock \emph{arXiv preprint arXiv:2010.09670}, 2020.

\end{thebibliography}
\clearpage
\appendix

\subsection{$\mathcal{A}$-distance}
We report the $\mathcal{A}$-distance values for all the $15$ corruptions in Table~\ref{app:tab:A-distance}
\begin{table*}[ht]
\centering
\begin{tabular}{@{}lrr@{}}
\toprule
 & \multicolumn{1}{l}{CIFAR10-C} & \multicolumn{1}{l}{Generated} \\ \midrule
gauss & 1.17 & 0.42 \\
shot & 1.12 & 0.43 \\
impulse & 1.36 & 0.55 \\
defocus & 0.95 & 0.92 \\
glass & 0.60 & 0.50 \\
motion & 0.75 & 0.65 \\
zoom & 0.79 & 0.73 \\
snow & 1.51 & 1.48 \\
frost & 1.39 & 1.36 \\
fog & 0.99 & 1.00 \\
brightness & 1.65 & 1.62 \\
contrast & 1.29 & 1.31 \\
elastic & 0.70 & 0.60 \\
pixelate & 1.21 & 0.85 \\
jpeg & 0.58 & 0.50 \\ 
mean & 1.07 & 0.86 \\ \bottomrule
\end{tabular}
\caption{We report the $\mathcal{A}$-distance between CIFAR10-C and CIFAR10 images and also between the images generated using our interpolated latent refinement scheme and CIFAR10 images}
\label{app:tab:A-distance}
\end{table*}

\subsection{Hyperparameter Details}
We perform $N = 1000$ steps of the diffusion process, use $\alpha = 0.9$ which is decided using the FID score of the generated images. During the model adaptation stage, we use a learning rate of $1e-3$, batch size of $200$, Adam optimizer with a $\beta = 0.9$ and we use $1$ optimizer step. We use a momentum of $0.999$ to update the teacher via exponential averaging.

 

\end{document}